\documentclass[conference]{IEEEtran}
\IEEEoverridecommandlockouts
\usepackage{cite}
\usepackage{amsmath,amssymb,amsfonts}
\usepackage{algorithmic}
\usepackage{graphicx}
\usepackage{textcomp}
\usepackage{xcolor}
\usepackage{slashbox}

\usepackage{tabularx}
\def\BibTeX{{\rm B\kern-.05em{\sc i\kern-.025em b}\kern-.08em
    T\kern-.1667em\lower.7ex\hbox{E}\kern-.125emX}}

\begin{document}

\title{Analyzing historical diagnosis code data from NIH N3C and RECOVER Programs using deep learning to determine risk factors for Long Covid\\
}

\makeatletter
\newcommand{\linebreakand}{%
  \end{@IEEEauthorhalign}
  \hfill\mbox{}\par
  \mbox{}\hfill\begin{@IEEEauthorhalign}
}

\makeatother

\author{\IEEEauthorblockN{Saurav Sengupta\IEEEauthorrefmark{1}, Johanna Loomba\IEEEauthorrefmark{2}, Suchetha Sharma\IEEEauthorrefmark{3}, Donald E. Brown\IEEEauthorrefmark{3}\IEEEauthorrefmark{1}}
\IEEEauthorblockN{Lorna Thorpe\IEEEauthorrefmark{4}, Melissa A Haendel\IEEEauthorrefmark{4}, Christopher G Chute\IEEEauthorrefmark{4}, Stephanie Hong\IEEEauthorrefmark{4} \\
on behalf of N3C and RECOVER consortium
}
\IEEEauthorblockA{\IEEEauthorrefmark{1}Department of Engineering Systems and Environment, University of Virginia}
\IEEEauthorblockA{\IEEEauthorrefmark{2}integrated Translational Health Research Institute of Virginia (iTHRIV), University of Virginia}
\IEEEauthorblockA{\IEEEauthorrefmark{3}School of Data Science, University of Virginia}
\IEEEauthorblockA{\IEEEauthorrefmark{4}NIH}
\{ss4yd, jjl4d, ss4jg, deb\}@virginia.edu, 

\IEEEauthorrefmark{4}lorna.thorpe@nyulangone.org, \IEEEauthorrefmark{4}melissa@tislab.org, \IEEEauthorrefmark{4} chute@jhu.edu, shong59@jh.edu
}

\maketitle

\begin{abstract}
Post-acute sequelae of SARS-CoV-2 infection (PASC) or Long COVID is an emerging medical condition that has been observed in several patients with a positive diagnosis for COVID-19. Historical Electronic Health Records (EHR) like diagnosis codes, lab results and clinical notes have been analyzed using deep learning and have been used to predict future clinical events. In this paper, we propose an interpretable deep learning approach to analyze historical diagnosis code data from the National COVID Cohort Collective (N3C)\footnote{https://ncats.nih.gov/n3c} to find the risk factors contributing to developing Long COVID. Using our deep learning approach, we are able to predict if a patient is suffering from Long COVID from a temporally ordered list of diagnosis codes up to 45 days post the first COVID positive test or diagnosis for each patient, with an accuracy of 70.48\%. We are then able to examine the trained model using Gradient-weighted Class Activation Mapping (GradCAM) to give each input diagnoses a score. The highest scored diagnosis were deemed to be the most important for making the correct prediction for a patient. We also propose a way to summarize these top diagnoses for each patient in our cohort and look at their temporal trends to determine which codes contribute towards a positive Long COVID diagnosis.

\end{abstract}

\begin{IEEEkeywords}
COVID-19, EHR, deep learning, GradCAM
\end{IEEEkeywords}

\section{Introduction}

Some people infected with the COVID-19 virus have demonstrated a wide range of health problems that can last a long time post infection, which has been termed Long COVID. According to the World Health Organization (WHO), approximately 10-20\% of the people infected with COVID-19 experience a variety of health conditions in the mid to long term after they recover from the initial illness. According to the NIH REsearching COVID to Enhance Recovery (RECOVER)\footnote{For more information on RECOVER, visit https://recovercovid.org/} program, which seeks to understand, treat, and prevent PASC, this condition generally refers to ongoing health effects, new or existing symptoms and other health problems that occur after the acute phase of SARS-Cov-2 infection (i.e., present four or more weeks after the acute infection). Therefore, it has become necessary to identify risk factors in a patient's medical history that can lead to them experiencing Long COVID.

The N3C repository contains records of patients with a newly introduced ICD-10\footnote{International Classification of Diseases 10th Revision} U09.9 code ("Post COVID-19 condition") that is being used to refer to patients being diagnosed with Long COVID \cite{haendel2021national}. The N3C repository also contains conditions, measurements and other medical records for these patients. We focus our analysis on all the recorded medical conditions recorded in the form of ICD-10 codes, for these patients up to 45 days post the first COVID diagnosis.

Previous efforts have focused on feature creation based on comorbidities, demographics, medication and healthcare utilization derived from the EHR data to develop machine learning models that can predict if a patient will develop Long Covid\cite{pfaff2022identifying}. In this work, we incorporate all conditions rather than limiting features to a pre-defined list of comorbidities to build a deep learning model to capture a more complete picture of a patient's medical history to find risk factors associated with Long COVID. 
We test different architectures for analyzing longitudinal data using all diagnosis codes present in a patient's medical history arranged temporally, and then use interpretability methods to identify conditions associated with the risk of developing Long COVID.

\section{Related Work}

Many previous works have analyzed temporal EHR data in different settings like predicting clinical events and risk stratification \cite{wang2019mcpl}. They have focused on Long Short Term Memory (LSTM) networks, a form of RNNs, to model longitudinal data 
\cite{ashfaq2019readmission}. 
RNNs are a class of neural network with looping connections between nodes such that temporal information persists \cite{hochreiter1997long} \cite{schuster1997bidirectional}. This makes them very useful to analyze time series data or applications where sequences have to be analyzed like speech recognition, language modeling etc. However, RNNs, by their nature, cannot remember long term dependencies in a sequence \cite{zhang2021recurrent}. LSTMs are a special kind of RNN that are architected to remember long term dependencies \cite{hochreiter1997long}. An LSTM unit consists of a cell and three gates: input, output and forget. The cell remembers information at each time step and the gates control the flow of information, that is, to either pass on or discard information to the next time step.









Zhang et al. \cite{zhang2019data} use an LSTM based model to generate representations for a cohort of Parkinson's disease patients. The input is a temporally ordered list of features $\{x_1, x_2 ... x_{N_p}\}$ at different times, extracted from the patient's EHR. A set of features are selected as prediction targets. At each time step $t_i$, the input is passed through two LSTM layers and the hidden state output of the final LSTM layer is used for calculating the loss functions.

An LSTM network is unidirectional, as in, it only preserves information from the past because it has only seen that part of the sequence. A bidirectional LSTM (BiLSTM), on the other hand, sees the input both ways,  that is, backwards (past) to forward (future) and forwards (future) to backward (past). Therefore, at any given time, we are able to use information from both the past and the future \cite{schuster1997bidirectional}. BiLSTM has been successfully used to analyze a patient's neuropsychological test scale data, genetic data and tomographic data in first, six and twelve months used to predict Alzheimer's Disease \cite{pan2019prediction}.

Attention has been used in addition to an LSTM network in \cite{suo2017multi}. It works by extracting the hidden states from an LSTM network and calculating an attention score $\alpha_t$ that can help weight the inputs by training an additional attention layer. We can later extract these attention weights during inference to understand which part of the input was given a higher weight during classification.

2D Convolutional Neural Networks (CNN) have been primarily used for computer vision applications, where multiple filters are trained to detect different input image features. 1D CNNs have been shown to work on time series problems like longitudinal EHR data \cite{ma2018risk}. 1D Convolution works over the temporal dimension with different filter sizes, where the different filters learn different temporal patterns. This process produces feature vectors which are then passed through a non-linear layer like a Rectified Linear Unit (ReLU) or Tanh.

Gradient-weighted Class Activation Mapping (GradCAM) has been used to examine 1D CNN based models that analyze protein sequences and find regions in the input sequences that help the model make the correct prediction \cite{kristianingsih2021accurate}. GradCAM is generally used in computer vision to generate localization maps for a given concept (class) in the input image \cite{selvaraju2017grad}. These maps are made by finding the gradient of the predicted class in the activation map of the final layer, pooling them channel wise, and the resultant activation channels are weighted with the corresponding gradients which can then be inspected to find which parts of the input helped in the classification.

Therefore, deep learning in general has been shown to provide value while analyzing clinical EHR data in a variety of areas. In the following sections, we define our methodology for analyzing the same in the N3C cohort.

\section{Methodology}

\subsection{Dataset}

The N3C data transfer to NCATS is performed under a Johns Hopkins University Reliance Protocol \# IRB00249128 or individual site agreements with NIH. The N3C repository contains N=14,026,265 number of  patients out of which X=5,409,269 are COVID-19 positive \cite{N3CCohor59}. COVID cases are defined as per CDC guidance \cite{centers2021coronavirus}. We construct our Long COVID positive cohort using patients with an existing U09.9 code or a long COVID clinic visit. The controls were constructed by choosing 5 random patients with the same site and within 90 days of the long COVID patient. At the end, we have 49,950 total patients, of which there were 7,511 Long COVID patients and 38,649 Control patients.

\subsection{Data pre-processing}

The N3C EHR repository contains all historical medical diagnosis codes stored using the Systematic Nomenclature of Medicine - Clinical Terms (SNOMED-CT) vocabulary for all patients. SNOMED-CT is a clinical terminology, widely used by healthcare providers for documentation and reporting within health systems \cite{agarwal2019snomed2vec}. Therefore, for each patient, we have a list of these diagnosis codes along with the date when the code was recorded. Since our goal is to find risk factors that can pre-dispose a patient to suffer from Long COVID, we focus on all conditions, not including the Long COVID diagnosis, in the patient's diagnostic history up to 45 days post the first COVID diagnosis or positive test, which we used as the acute phase cut off. We arrange all these conditions in an ordered list from the earliest to the latest. We also ensure that we insert only one record in the ordered list for all conditions that were repeatedly recorded in a single day, which can occur when one patient can have multiple tests or diagnoses in a single day. At the end of this process, each patient $p_i$ has an ordered list of diagnosis codes $[d_1, d_2, ...d_K]$, where $K =1000$. We select $K=1000$ as the upper limit for the length of the list of diagnosis codes as we found that 99\% of our patients had less than 1000 diagnosis codes present in their medical history. We add padding using "padding tokens" ($[PAD]$) to make all inputs of length shorter than 1000 of uniform length. For those conditions for which we do not have prior embeddings, we replace with $[UNK]$ tokens.

\subsection{Pre-trained SNOMED-CT embeddings}

Prior work focuses on embedded vector representation learning to make medical concepts analyzable via mathematical models and subsequently building models for analysis \cite{wang2019mcpl}. To analyze these temporal patterns in an ordered list of concept codes using deep learning, we first have to transform them into their equivalent vector representations that also capture semantic meaning and similarities between different diagnoses. We used 200-dimensional SNOMED embeddings trained using SNOMED2Vec, a graph based representation learning method on SNOMED-CT, to generate meaningful representations \cite{agarwal2019snomed2vec}. This will help us capture more meaning in these embeddings than randomly initializing the embedding layer and training from scratch. In effect, for each diagnosis code, $d_i$ we have a representation vector, $v_i$. For $[PAD]>$ tokens, we initialize the embeddings as all 0s and for $[UNK]$ tokens, we initialize them as the mean of all embeddings.

\subsection{One level roll-up in SNOMED-CT hierarchy}
Due to the recent nature of the disease, we do not have a pre-trained embedding for the "COVID-19" diagnosis code. Being the reason for Long COVID, we need to capture that information in our analysis. Therefore, we do a one level roll-up in the SNOMED hierarchy for "Coronavirus Infection" for which we do have a pre-trained embedding provided in \cite{agarwal2019snomed2vec}. We do this roll-up for all diagnosis codes in our dataset to prevent mixing hierarchical information. Since "Coronavirus Infection" encompasses all the different forms of the virus, we remove all such instances in a patient's record and insert only the first recorded positive COVID-19 infection in the temporally ordered list. Since a patient could have more than one COVID test or diagnosis, we remove the repetitions and insert only the earliest COVID diagnosis or test in the ordered list. We term the date of the earliest COVID diagnosis or test as the \textbf{COVID index date}.




\subsection{Modeling}

\begin{figure}[htbp]
\centerline
{\includegraphics[width=0.5\textwidth]{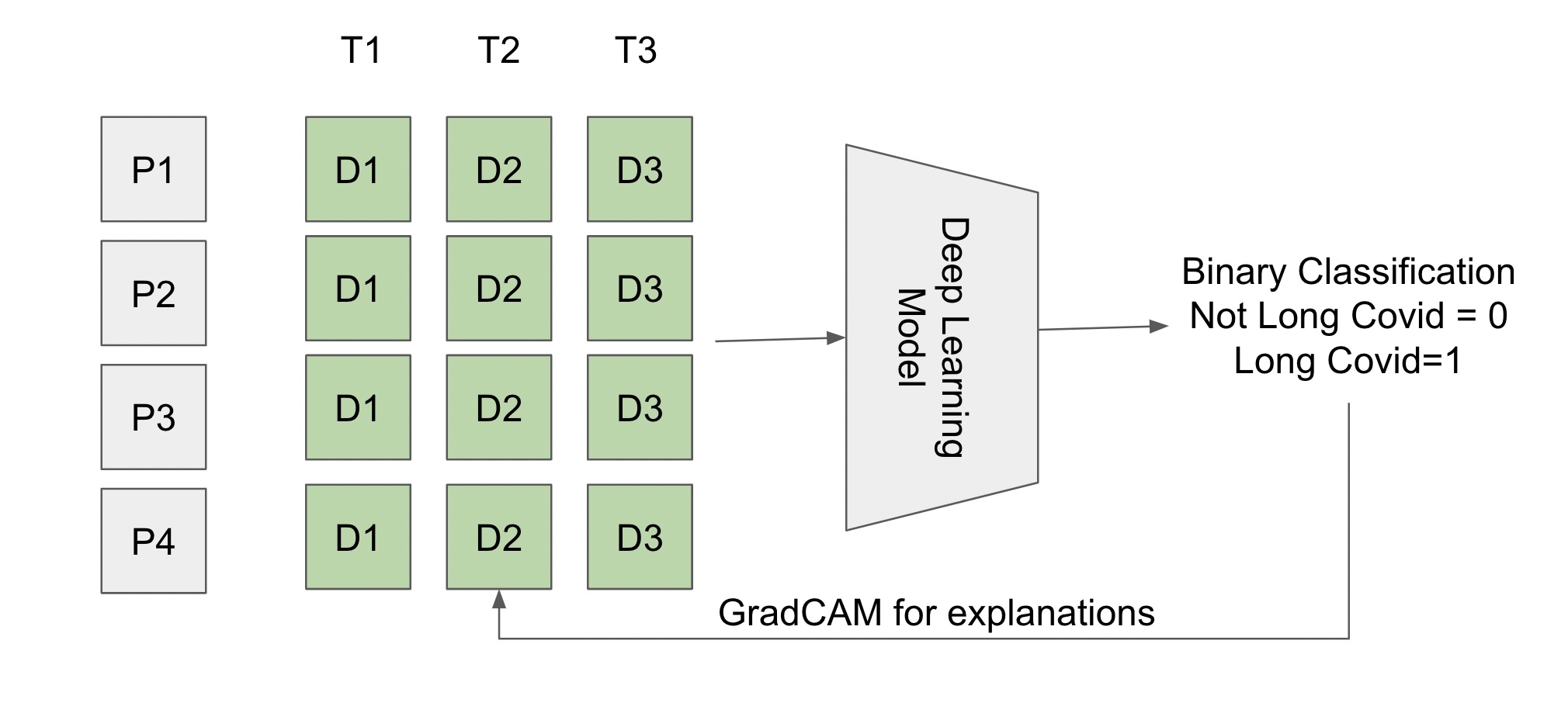}}
\caption{Overview of training. Patients $P_i$ have diagnoses $D_i$ at times $T_i$.}
\label{fig}
\end{figure}

Here we provide the details of the different architectures we used to analyze our data. The size of the LSTM refers to the dimensions of the hidden layer as defined in the PyTorch documentation. For each model tested, we initialized the embedding layer using the 200-dimensional pre-trained SNOMED-CT embeddings provided by \cite{agarwal2019snomed2vec}, and froze it during training to preserve the embeddings.
At each training epoch, all patient input sequences $[d_1, d_2, ... d_K]$ are passed through the embedding layer, and subsequent layers defined below, and the cross entropy loss is evaluated against target Long COVID labels for those patients.

\subsubsection{2 layer unidirectional LSTM}

We used a 2 layer LSTM each of size 128 and then added an output linear layer to build a binary classification model.

\subsubsection{2 Layer bidirectional LSTM}

We used a 2 layer bidirectional LSTM each of size 128 and then added an output linear layer to build a binary classification model.

\subsubsection{2 Layer Bidirectional LSTM with attention layer}

We use a self attention layer in between the LSTM layer and the output linear layer, drawing inspiration from \cite{suo2017multi}. 
Given a hidden layer representation from the LSTM, $h_t$, we calculate attention using the following equations:

\begin{equation}
    \begin{aligned}
        u_t = tanh(Wh_t + b) \\
        \alpha_t = softmax({v}^Tu_t) \\
        s = \sum_{t=1}^{M} \alpha_t h_t
    \end{aligned}
\end{equation}

where $\alpha_t$ are the attention weights that help the network focus on specific parts of the input to generate the correct output, $M$ is the length of the input sentence and $v_t$ is the learned vector during training. $u_t$ could be thought of as a non-linear projection of $h_t$.

\subsubsection{2 Layer Bidirectional LSTM with a 1D CNN unit}

We add a convolution unit consisting o a 1D-CNN layer, a batch norm layer and ReLU non-linearity layer. The CNN outputs 256 dimensional features and the kernel size is 3. We also add a max pool layer of size 2 to aggregate the feature maps, between the output and LSTM layers (Figure \ref{bilstm}).

\begin{figure}[htbp]
\centerline
{\includegraphics[width=0.5\textwidth]{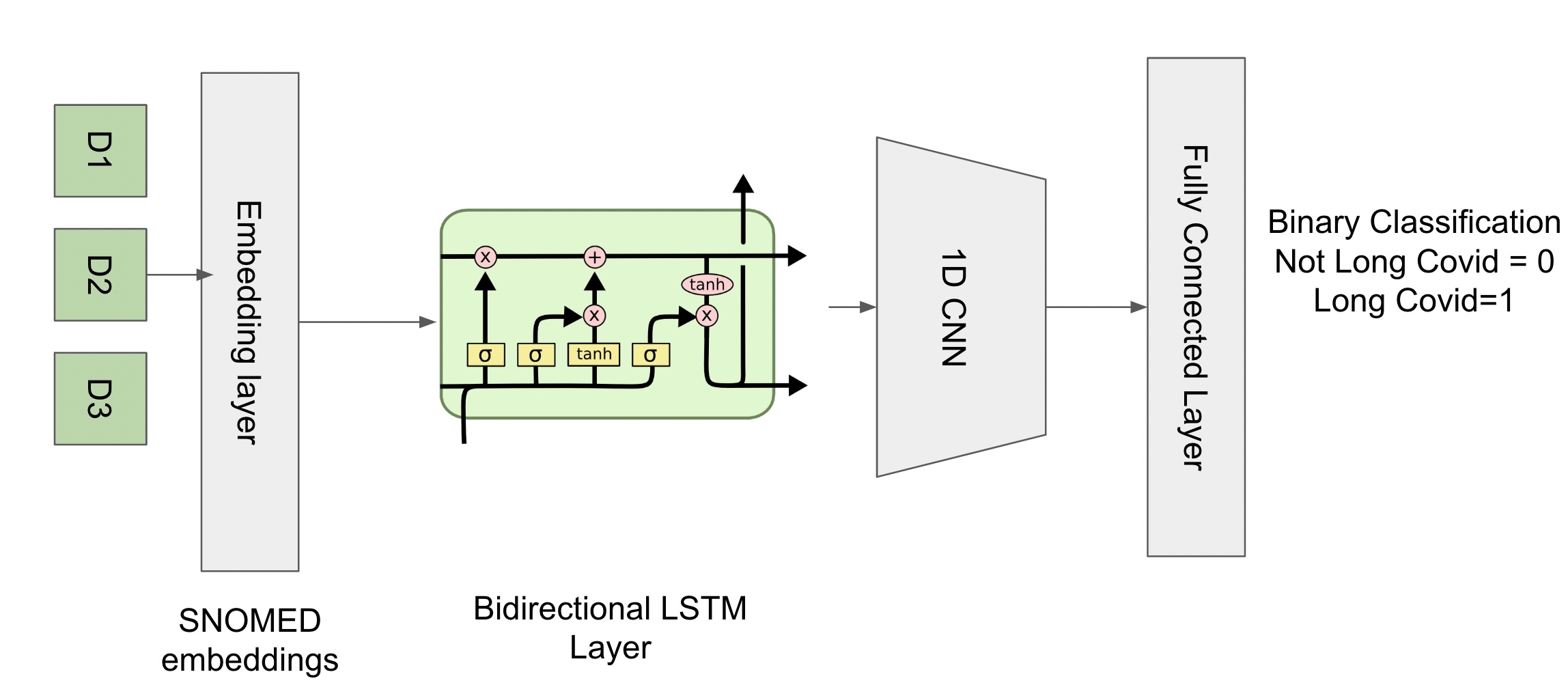}}
\caption{BiLSTM CNN architecture.}
\label{bilstm}
\end{figure}

\begin{figure*}[t]
\centering
\includegraphics[width=\textwidth]{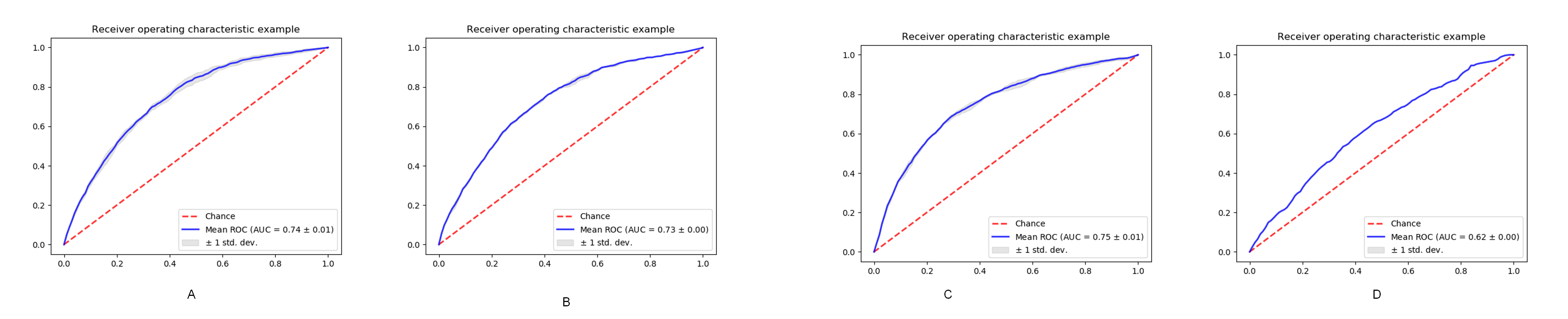}
\caption{ROC Curves for different architectures tested. A: 2 layer Unidirectional LSTM, B: 2 Layer Bidirectional LSTM, C: 2 Layer Bidirectional LSTM with 1D CNN, D: 2 Layer Bidirectional LSTM with attention layer}
\label{roc_curves}
\end{figure*}

\subsection{Training details}

We split the data in training, validation and testing in the ratio of 75:15:10 and use a batch size of 64. Since we are doing binary classification, we label Long COVID patients as 1 and the Controls as 0 and use the cross entropy loss function to calculate the loss. For training, we use Adadelta optimization algorithm with a learning rate of 0.01. We decay the learning rate by a factor of 0.8 if the validation loss does not decrease for 8 consecutive epochs. RNNs can suffer from the exploding gradient problem, in which the updates to the weights can be very large due to multiple updates during the time steps for which we use gradient clipping, where we specify lower and upper bounds for a gradient \cite{zhang2019gradient}. We set the bounds as (-5,5) during training.



\subsection{GradCAM calculation and analysis}
We calculate the GradCAM heatmap values by first calculating the gradients from the conv\_1d layer with respect to the correct prediction of Long COVID from the forward pass of the model \cite{selvaraju2017grad}. We extract the activation map from the CNN layer and weight them using the pooled gradients and average over all channels. We pass these through a Rectified Linear Unit (ReLU) function and normalize the values using max scaling to generate the scores for each input diagnosis.

We use GradCAM to assign scores to each input diagnosis code for each patient that our model correctly predicts as having Long COVID. We then pick the diagnosis code with the highest GradCAM score and calculate the time separation (in days) from when the diagnosis code was recorded to the COVID index date. We do this for all patients and capture the distribution of the time separation to understand when the highest scored diagnosis was made across different patients at different sites.

\begin{figure*}[t]
\centering
\includegraphics[width=\textwidth]{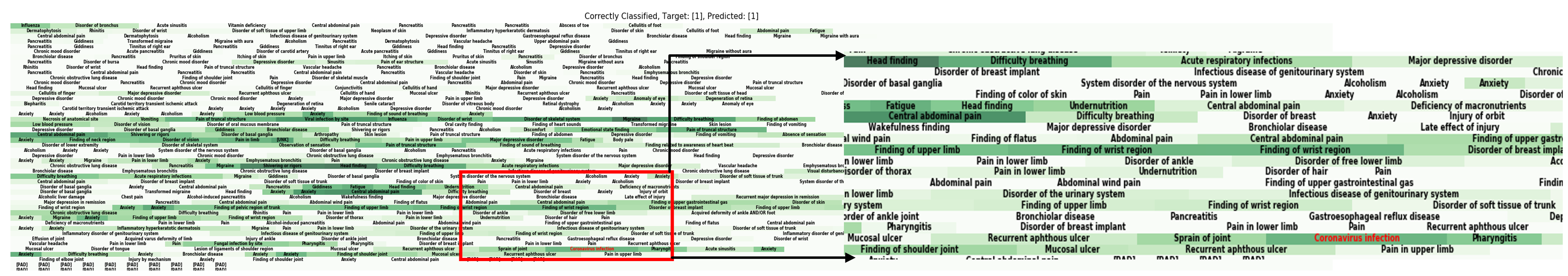}
\caption{GradCAM visualization of correctly classified Long COVID patients.}
\label{gradcam_single}
\end{figure*}

\section{Results}

Since the dataset is imbalanced, we use Area Under the Receiver Operating Characteristic (AUC) curve to determine the effectiveness of the model. We present the mean AUC scores of the 3-fold stratified cross validation for all models tested in Table \ref{meanauc}. Refer to Figure \ref{roc_curves} for ROC curves. We choose the model with the highest test AUC and calculate the best classification threshold using Youden's J statistic \cite{ruopp2008youden}. This results in a 70.48 \% classification accuracy in our test set for the fold with the best AUC, which was 0.75.

\begin{table}[htbp]
\caption{Results}
\begin{center}
\begin{tabular}{|p{4.7cm}|p{3cm}|}
\hline
\textbf{Model}&{\textbf{Mean AUC (3-fold CV)}} \\
\hline
Unidirectional LSTM & 0.74 $\pm$ 0.01 \\
\hline
Bidirectional LSTM& 0.73 $\pm$ 0.01\\
\hline
Bidirectional LSTM with 1D CNN unit& \textbf{0.75 $\pm$ 0.01}\\
\hline
Bidirectional LSTM with attention layer& 0.62 $\pm$ 0.00\\
\hline
\end{tabular}
\label{meanauc}
\end{center}
\end{table}





\section{Analysis}


Based on the results in Table \ref{meanauc}, we pick the model with the highest mean AUC, which is the BiLSTM with 1D CNN model. We calculate the GradCAM scores for each input condition for a \textbf{correctly} classified Long COVID patient and look at the condition which received the highest score. We also find the time separation (in days) of the highest score condition from the COVID index date. If Condition $A$ occurred $t$ days before the index date, the time separation is noted as $-t$, and $+t$ if it occurred $t$ days after. We then look at the distribution of these diagnosis codes over all correctly predicted patients in our cohort (see Figure \ref{dist}). In Figure \ref{gradcam_single} we present a visualization of the GradCAM values of a correctly classified patient's diagnosis history with darker shades representing higher values. This can help a physician determine what the model thought of as important for making its classification. We can use this to investigate the model and provide a guidance to physicians that which diagnoses in a patient's history to take a closer look at. We defer a more detailed clinical analysis for subsequent papers.

Hence, our proposed method is able to analyze \textbf{all} historical patient diagnostic data and provide interpretable results that can be used to identify risk factors that can contribute (pending further clinical analysis) to a Long COVID diagnosis. 
\begin{figure}[htbp]
\centerline
{\includegraphics[width=0.4\textwidth]{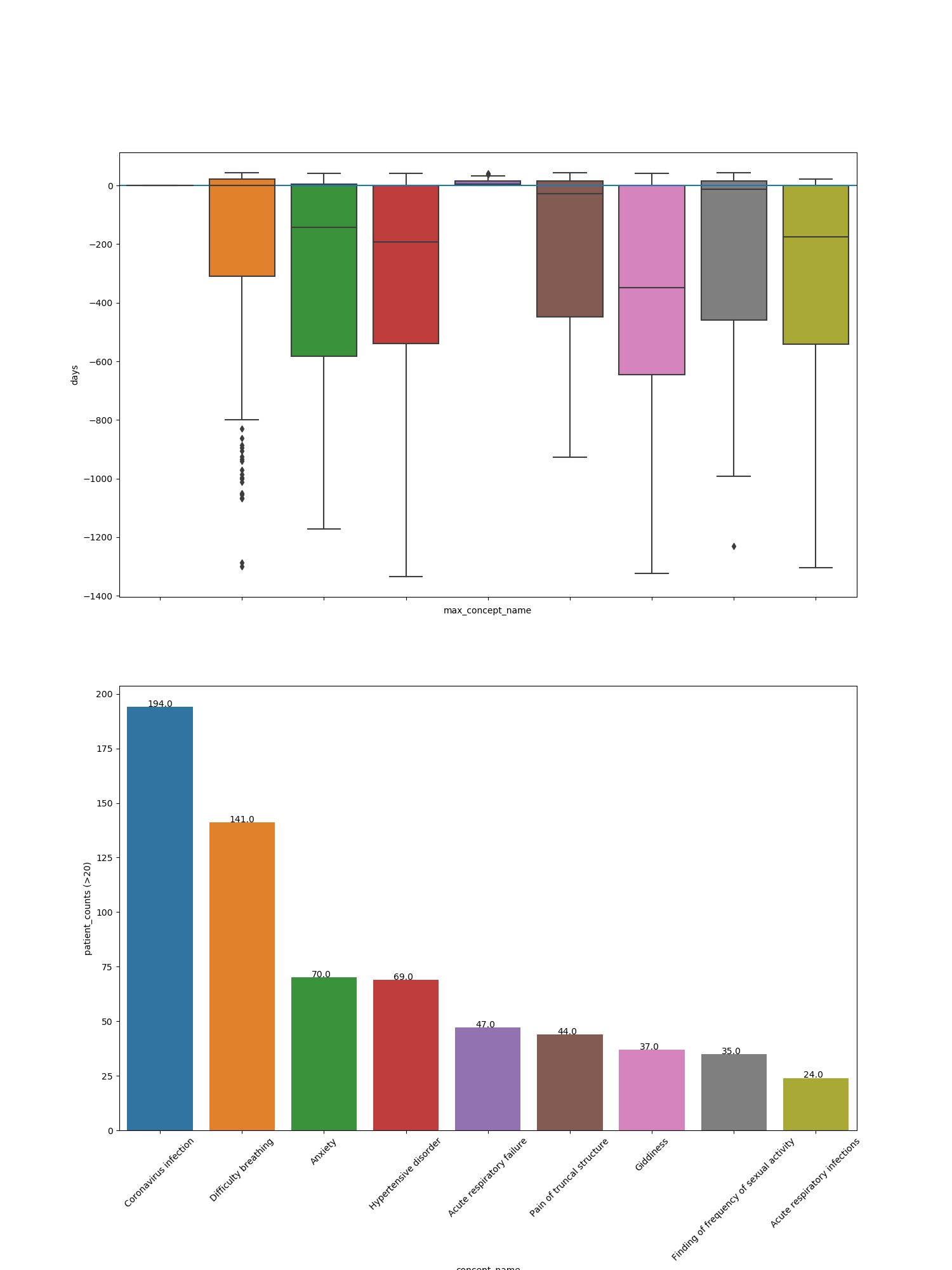}}
\caption{Distribution of time separation (in days) of the top 9 (by number of patients) diagnosis codes over all correctly predicted patients from the COVID index date (0-line).}
\label{dist}
\end{figure}

\section{Limitations and Future Work}

While we define our Control patients as those without a U09.9 code or Long COVID clinic visit, we cannot be sure that those patients were definitely not suffering from Long COVID. This introduces noise in the dataset. We defer a more involved clinical analysis for future work as the focus of this paper is to summarize the methodology used to identify historical conditions in N3C data.
Future work would focus on Positive Unlabeled or PU learning, where we know our positive labels, but our negative labels are unknown \cite{su2021positive}.
Future efforts will be made to incorporate more recent deep learning approaches for uneven temporal sequences, like self-attention based Transformers, to analyze this data \cite{li2020behrt}. 
Secondary use of medical record data is always subject to bias, as patient events are only represented as captured in medical records of participating sites. N3C data is heterogeneous across contributing sites due to variation in local care practices, EHR encoding, and clinical data model mapping. We randomized our data partitioning, but further work can be done to investigate the effect of site data heterogeneity. Hence, this analysis does not reflect a representative sample of all patients, but rather a subset of facts collected by participating academic medical institutions.

\section*{Acknowledgment}

The analyses described in this publication were conducted with data or tools accessed through the NCATS N3C Data Enclave covid.cd2h.org/enclave and supported by CD2H - The National COVID Cohort Collaborative (N3C) IDeA CTR Collaboration 3U24TR002306-04S2 NCATS U24 TR002306. This research was possible because of the patients whose information is included within the data from participating organizations (covid.cd2h.org/dtas) and the organizations and scientists (covid.cd2h.org/duas) who have contributed to the ongoing development of this community resource \cite{haendel2021national}. The content is solely the responsibility of the authors and does not necessarily represent the official views of the RECOVER Program, the NIH or other funders. We would like to thank the National Community Engagement Group (NCEG), all patient, caregiver and community Representatives, and all the participants enrolled in the RECOVER Initiative.
We gratefully acknowledge the core contributors to N3C present here, covid.cd2h.org/core-contributors. Authorship was determined using ICMJE recommendations.

\section*{Funding Statement}

This research was funded by the National Institutes of Health (NIH) Agreement OT2HL161847-01. The views and conclusions contained in this document are those of the authors and should not be interpreted as representing the official policies, either expressed or implied, of the NIH. All authors are funded under this mechanism.
Also funded by the NIH award UL1TR003015 for the integrated Translational Health Research Institute of Virginia.







\bibliographystyle{IEEEtran}
\bibliography{references.bib}

\vspace{12pt}


\end{document}